\documentclass[journal]{IEEEtran}

\usepackage{cite}
\usepackage{amsmath,amssymb,amsfonts}
\usepackage{algorithm}
\usepackage{algpseudocode}
\usepackage{graphicx}
\usepackage{textcomp}
\usepackage{multicol}
\usepackage{cuted}
\usepackage{extarrows}
\usepackage{booktabs}  % For professional-looking tables
\usepackage{multirow}
\usepackage[table]{xcolor}
\usepackage{lipsum}
\usepackage{enumitem}

\usepackage{array}

\definecolor{headercolor}{rgb}{0.369, 0.788, 0.478}
\definecolor{positivecolor}{rgb}{0.494, 0.753, 0.933}
\definecolor{negativecolor}{rgb}{0.988, 0.553, 0.384}

\usepackage{amsthm}
\theoremstyle{definition}

\usepackage{multirow}
\usepackage{subcaption}

\usepackage{nicematrix}

\def\BibTeX{{\rm B\kern-.05em{\sc i\kern-.025em b}\kern-.08em
    T\kern-.1667em\lower.7ex\hbox{E}\kern-.125emX}}
    
\IEEEoverridecommandlockouts
\begin{document}

% Old Title: Cryptocurrency Price Forecasting Using XGBoost Regression: A Machine Learning Approach 

\title{Cryptocurrency Price Forecasting Using XGBoost Regressor and Technical Indicators} 

\author{Abdelatif Hafid$^1$, Maad Ebrahim$^2$, Ali Alfatemi$^3$, Mohamed Rahouti$^3$, and Diogo Oliveira$^4$

\textit{$^1$ESISA Analytica, Higher School of Engineering in Applied Sciences, Fez, Morocco}

\texttt{abdelatif.hafid@yahoo.com}

\textit{$^2$GAIA, Ericsson, Montreal, Canada} 

\texttt{maad.ebrahim@ericsson.com}

\textit{$^3$Computer and Information Science, Fordham University, NY 10023, USA.} 

\texttt{mrahouti@fordham.edu; aalfatemi@fordham.edu}

\textit{$^4$College of IST, Penn State University, Monaca, PA 15601, USA.} 

\texttt{dko5179@psu.edu} 

%\texttt{aalfatemi@fordham.edu}

% \textit{$^3$Department of Statistics, University of PQR, City, Country}}
}

% Please add my PhD student Ali: 
%\IEEEauthorblockN{Ali Alfatemi}
    %\IEEEauthorblockA{\textit{Computer \& Information Science Department} \\
    %\textit{Fordham University}\\
    %Bronx, NY 10458, USA \\
    %aalfatemi@fordham.edu}

\maketitle

\begin{abstract}
The rapid growth of the stock market has attracted many investors due to its potential for significant profits. However, predicting stock prices accurately is difficult because financial markets are complex and constantly changing. This is especially true for the cryptocurrency market, which is known for its extreme volatility, making it challenging for traders and investors to make wise and profitable decisions. This study introduces a machine learning approach to predict cryptocurrency prices. Specifically, we make use of important technical indicators such as Exponential Moving Average (EMA) and Moving Average Convergence Divergence (MACD) to train and feed the XGBoost regressor model. We demonstrate our approach through an analysis focusing on the closing prices of Bitcoin cryptocurrency. We evaluate the model's performance through various simulations, showing promising results that suggest its usefulness in aiding/guiding cryptocurrency traders and investors in dynamic market conditions. 
\end{abstract}


\section{Introduction} \label{sec: introduction} 

Over the past few years, the rapid expansion of the stock market has made it an appealing option for investors seeking high returns and easy access. However, investing in stocks carries inherent risks, underscoring the need for a well-defined investment strategy. Traditionally, investors relied on empirical methods such as technical analysis, guided by financial expertise. With the widespread adoption of financial technology (FinTech), statistical models incorporating machine learning techniques have emerged for forecasting stock price movements. This shift has demonstrated significant success across various markets, including the S\&P 500, NASDAQ \cite{hsu2021fingat}, and the cryptocurrency market \cite{liu2021forecasting, hafid2023bitcoin}. In this research, our emphasis is on the cryptocurrency market, a dynamic force in finance, with a particular focus on Bitcoin price prediction \cite{nakamoto2008bitcoin}.

Furthermore, Blockchain technology, the backbone of cryptocurrencies, has gained substantial attention in the banking and financial industry due to its secure and transparent decentralized database \cite{hafid2020scaling}. Despite the advantages of abundant market data and continuous trading, the cryptocurrency market faces challenges such as high price volatility and relatively smaller capitalization. Success in cryptocurrency financial trading hinges on the careful analysis and selection of data, making the development of machine learning models crucial for extracting meaningful insights. Models such as Long Short Term Memory (LSTM) and Random Forest (RF) are instrumental in predicting cryptocurrency prices by leveraging historical data and patterns, thereby aiding effective decision-making in this volatile market. Despite the potential, there have been limited studies attempting to create successful trading strategies in the cryptocurrency market.

With the advent of FinTech, machine learning models have been increasingly adopted to forecast stock price movements, transforming the landscape of financial analysis and trading. These models leverage large datasets and complex algorithms to identify patterns and predict future price trends, which has led to notable success across various markets, including the S\&P 500 and NASDAQ \cite{hsu2021fingat}. In the cryptocurrency market, which is characterized by its high volatility and rapid price fluctuations, machine learning techniques have proven particularly valuable. Studies have demonstrated the efficacy of deep learning methods, such as Stacked Denoising Autoencoders (SDAE) and LSTM networks, in predicting Bitcoin prices with high accuracy \cite{liu2021forecasting, hafid2023bitcoin}. These models utilize a variety of inputs, including historical price data, trading volume, public sentiment, and macroeconomic indicators, to generate predictions that can guide investment decisions. The integration of machine learning into FinTech has thus provided investors with powerful tools to navigate the complexities of financial markets, enhancing their ability to make informed and strategic trading decisions.

Despite the advantages of the cryptocurrency market, such as abundant market data and continuous trading, it faces significant challenges like high price volatility and relatively smaller capitalization. Successful trading in this market depends on careful data analysis and selection, making the development of machine learning models crucial for extracting meaningful insights. Models like LSTM and RF are instrumental in predicting cryptocurrency prices by utilizing historical data and patterns, thus aiding effective decision-making in this volatile landscape. While there have been limited studies on developing successful trading strategies in the cryptocurrency market, our research aims to bridge this gap by introducing a novel machine learning strategy using the XGBoost regressor model, which incorporates essential technical indicators and historical data to enhance financial trading strategies.

%\textcolor{red}{This research focuses on the cryptocurrency market, particularly Bitcoin price prediction \cite{nakamoto2008bitcoin}, leveraging blockchain technology which serves as a secure and transparent decentralized database \cite{hafid2020scaling}.}

This research introduces an efficient machine learning approach for forecasting cryptocurrency prices, specifically focusing on Bitcoin. The motivation behind this study stems from the inherent volatility and complexity of the cryptocurrency market, which pose significant challenges for traders and investors. Traditional methods of technical analysis and empirical strategies are often insufficient in predicting price movements in such a dynamic environment. To address this, we propose using the XGBoost regressor model, a powerful machine learning technique known for its robustness and accuracy. Our methodology integrates a comprehensive set of technical indicators, including the Exponential Moving Average (EMA), Moving Average Convergence Divergence (MACD), Relative Strength Index (RSI), and other relevant metrics derived from historical market data. The data is sourced from Binance via its API, covering a detailed time span with high-frequency intervals, which allows for capturing rapid market changes.

The proposed model undergoes extensive preprocessing and feature engineering to enhance its predictive capabilities. By employing regularization techniques, we mitigate the risk of overfitting and fine-tune the model parameters through a grid search for optimal performance. Our results demonstrate that the XGBoost regressor model significantly improves prediction accuracy, evidenced by low Mean Absolute Error (MAE) and Root Mean Squared Error (RMSE) values, as well as a near-perfect R-squared value. This study contributes to the state-of-the-art by providing a robust and scalable solution for cryptocurrency price prediction, leveraging advanced machine learning techniques to navigate the complexities of financial markets and aiding in informed decision-making for traders and investors.

The key contributions of this paper are summarized as follows:
\begin{itemize}
    \item Introduce an efficient machine learning strategy using the XGBoost regressor model for cryptocurrency price prediction.
    \item Integrate a comprehensive set of technical indicators, including EMA, MACD, RSI, and others, with historical market data.
    \item Employ regularization techniques to mitigate overfitting and fine-tuned model parameters through grid search.
    \item Demonstrate significant improvements in prediction accuracy with low MAE, RMSE, and a near-perfect R-squared value.
    \item Provide a robust and scalable solution for navigating the complexities of financial markets, aiding informed decision-making for traders and investors.
\end{itemize}

The rest of the paper is organized as follows. Section \ref{sec:related_work} reviews the most relevant existing works. Section \ref{sec:data} explains how we collected and prepared the data. Section \ref{sec:mathematical_model} proposes the machine learning model and its mathematical formulation. In Section \ref{sec:results_and_analysis}, we evaluate/assess our proposed model. Section \ref{sec:results_and_analysis} also provides a comparison between the proposed work and existing studies in the literature. Finally, Section \ref{sec:conclusion} concludes the paper.

\section{Related Work} 
\label{sec:related_work}
%\textcolor{red}{mrahouti: A section for related work/state-of-the-art discussion is needed. I can help put some material here.}

%\textcolor{red}{The rapid growth and extreme volatility of the cryptocurrency market have driven significant interest in developing accurate price forecasting models. This section reviews the state-of-the-art studies employing machine learning techniques to predict cryptocurrency prices, particularly Bitcoin, due to its market dominance and extensive data availability. These studies illustrate the advancements and applications of various machine learning methods, highlighting their effectiveness in navigating the complexities of digital currency markets.}

The effort to forecast cryptocurrency prices has garnered significant interest in recent years, leading to the development of various methods to address this complex problem \cite{sebastiao2021forecasting}. This section reviews advanced studies that employ machine learning for predicting cryptocurrency prices, with a particular focus on Bitcoin due to its dominant position and the extensive availability of data.

Among these advancements, machine learning has significantly impacted cryptocurrency price forecasting by providing models that adeptly navigate the complex and volatile digital currency market \cite{khedr2021cryptocurrency}. These methods range from simple regression models to advanced deep learning networks, each capable of detecting patterns and predicting future prices based on historical data \cite{nazareth2023financial}.

Cryptocurrency value fluctuations are influenced by numerous factors, which has prompted the adoption of machine learning for price prediction \cite{tanwar2019machine, awotunde2021machine}. For instance, studies by Greaves and AU \cite{1:IEEEtranpage} have investigated using network attributes and machine learning to predict Bitcoin prices. Similarly, Jang and Lee \cite{jang2017empirical} combined blockchain-related features, time series analysis, and Bayesian neural networks (BNNs) for Bitcoin price analysis.

Building on this foundation, further research by \cite{mcnally2018predicting}, \cite{velankar2018bitcoin}, and \cite{saad2019toward} has applied machine learning to Bitcoin price forecasting. Saad et al. \cite{saad2019toward} not only predicted prices but also identified critical network attributes and user behaviors influencing price variations in Bitcoin and Ethereum \cite{wood2014ethereum}, alongside the supply and demand dynamics of cryptocurrencies. Additionally, Sin and Wang \cite{sin2017bitcoin} utilized neural networks for price predictions, leveraging blockchain data features.

Continuing this trend, Christoforou et al. \cite{christoforou2020neural} developed a Bitcoin price prediction model using neural networks, focusing on factors affecting price volatility and utilizing blockchain data and network activity metrics for forecasting. Furthermore, Chen et al. \cite{chen2020bitcoin} and Akyildirim et al. \cite{akyildirim2023forecasting} demonstrated the application of machine learning in forecasting Bitcoin prices and mid-price movement of Bitcoin futures, respectively. These studies highlight the ability of machine learning to harness vast datasets and identify complex patterns, enhancing predictive accuracy beyond traditional statistical approaches.

Moreover, some studies have demonstrated the effectiveness of combining machine learning techniques with blockchain data for cryptocurrency price forecasting. For example, Martin et al. \cite{martin2020combining} introduced a hybrid method that merges diverse data and analytical techniques, enhancing accuracy in this complex field. Liu et al. \cite{liu2023financial} focused on optimizing performance and interpretability in financial time series, showcasing the benefits of combining various machine learning approaches. He et al. \cite{he2023financial} developed a deep learning ensemble model for financial time series forecasting, applicable to cryptocurrencies, illustrating the increased reliability and accuracy of multiple deep learning \cite{alfatemi2024advancing} strategies.

\begin{table*}[ht!]
\renewcommand{\arraystretch}{1.1}
\caption{Notations and abbreviations used in the paper.}
\label{tab:notation}
\centering
\small
\begin{tabular}{lp{0.7\linewidth}}
\hline
\textbf{Notation} & \textbf{Description} \\
\hline
\(m\) & Total number of samples or observations\\
\(m_{\text{train}}\) & Number of training samples \\
\(m_{\text{test}}\) & Number of testing samples \\
$n$ & Number of features \\
$\mathcal{C}_p^{(i)}$ & Close price at time $t_i$ \\
$\mathcal{O}_p^{(i)}$ & Opening price at time $t_i$ \\
$\mathcal{H}_p^{(i)}$ & High price at time $t_i$\\
$\mathcal{L}_p^{(i)}$ & Low price at time $t_i$ \\
$\mathcal{V}^{(i)}$ & Volume of the cryptocurrency being traded at time $t_i$ \\
$\text{QAV}^{(i)}$ & Total trading value at time $t_i$  \\
$\text{NOT}^{(i)}$ & Number of trades at time $t_i$  \\
$\text{TBBV}^{(i)}$ & Total volume of Bitcoin bought at time $t_i$  \\
$\text{RSI}_{\alpha}^{(i)}$ & Relative strength index at time $t_i$ within a time period $\alpha$ \\
$\text{MACD}^{(i)}$ & Moving average convergence divergence at time $t_i$ \\
$\text{EMA}_{\alpha}^{(i)}$ & Exponential moving average at time $t_i$ within a period of time $\alpha$ \\
$\text{PROC}_{\alpha}^{(i)}$ & Price rate of change at time $t_i$ within a period of time $\alpha$\\
$ \%K_{\alpha}^{(i)}$  & Stochastic oscillator at time $t_i$ within a period of time $\alpha$ \\ 
$\text{MOM}_{\alpha}^{(i)}$ & Momentum at time $t_i$ within a period of time $\alpha$ \\ 
$\eta$ & Learning rate \\
$\lambda$, $\alpha$ & Regularization parameters \\
$N$ & Number of trees  \\
$\Delta t$ & Time interval \\
\hline
\end{tabular} 
\normalsize
\end{table*}

\begin{figure}[!t]
   \centering
   \includegraphics[width=3.5in]{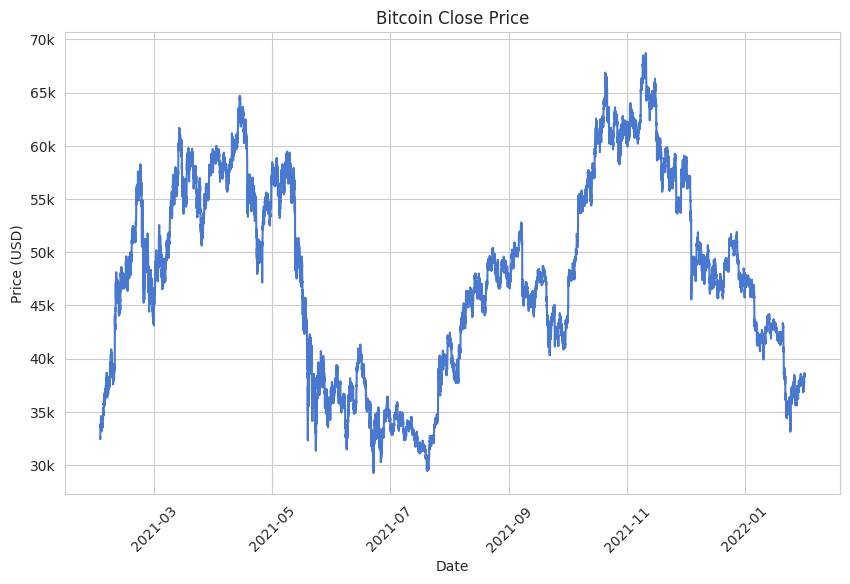}
   \caption{Bitcoin close price over time.}
   \label{fig:bitcoin_close_price}
\end{figure}

Additionally, Nazareth and Reddy \cite{nazareth2023financial} reviewed machine learning in finance, highlighting hybrid models' effectiveness in handling financial market complexities. Further research by Nagula and Alexakis \cite{nagula2022new}, Petrovic et al. \cite{petrovic2021cryptocurrency}, Gupta and Nalavade \cite{gupta2023metaheuristic}, and Luo et al. \cite{luo2022bitcoin} underscores the success of diverse computational techniques in improving Bitcoin price predictions, advancing sophisticated, accurate models for cryptocurrency investments.

In conclusion, machine learning not only excels in predictive accuracy but also in adaptability and scalability, both of which are essential as the cryptocurrency market evolves. With the capacity to update models with new data, machine learning remains a vital tool for cryptocurrency trading and investment, ensuring timely and precise forecasts \cite{chen2020bitcoin, akyildirim2023forecasting}.

Unlike existing studies, our work introduces a novel machine learning strategy that leverages the XGBoost regressor model, combining a range of technical indicators such as EMA and MACD with historical data for Bitcoin price prediction. This approach emphasizes the use of regularization techniques to prevent overfitting and fine-tuning model parameters for enhanced accuracy. Our methodology stands out by effectively integrating diverse datasets and analytical techniques, ensuring robust and precise predictions in the highly volatile cryptocurrency market.

% ---- Data ----
% ----
\section{Data}
\label{sec:data}
% In this section, we cover the essential notations and abbreviations, explain how we collected the data, detail our process for preprocessing the data, and eventually discuss how we engineered additional features. Table \ref{tab:notation} provides the definitions of the parameters and abbreviations used in this paper.

This section covers the essential notations and abbreviations, explains the data collection process, details the preprocessing steps, and discusses the engineering of additional features. Table \ref{tab:notation} provides the definitions of the parameters and abbreviations used in this paper.

\subsection{Data Collection and Preprocessing}\label{subsec:data_collection _and_preprocessing}

%We obtain Bitcoin historical market data from Binance via Binance API \cite{binance_data}. The dataset covers the time period from February 1, 2021, to February 1, 2022, with a time interval of 15 minutes (i.e., $\Delta t$ = $15$ minutes). We choose $15$ minutes because it gives us good accuracy. We split the data into $80\%$ for training set and $20\%$ for testing set.

%We choose this shorter time interval for several reasons, including the high volatility of the Bitcoin market. In highly volatile markets, shorter intervals are often preferred to capture rapid changes, whereas less volatile markets may be suitable for longer intervals. 

\begin{algorithm*}
\small
\caption{Cryptocurrency price forecasting using XGBoost regressor and technical indicators.}
\label{alg:crypto_forecasting}
\begin{algorithmic}[1]
\State \textbf{Input}: Historical market data $H = \{h_1, h_2, ..., h_T\}$, Technical indicators $T = \{t_1, t_2, ..., t_N\}$, Target variable $Y = \{y_1, y_2, ..., y_T\}$
\State \textbf{Output}: Trained XGBoost regressor model $\mathcal{M}$

\Procedure{Data Preparation}{}
    \State \textit{// Collect and preprocess data}
    \State $D \leftarrow \text{Collect data from Binance API}$
    \State $D_{\text{train}}, D_{\text{test}} \leftarrow \text{Split data into training and testing}$
    \State \textit{// Scale features}
    \State $D_{\text{train}} \leftarrow \text{StandardScaler.fit\_transform}(D_{\text{train}})$
    \State $D_{\text{test}} \leftarrow \text{StandardScaler.transform}(D_{\text{test}})$
\EndProcedure

\Procedure{Feature Engineering}{}
    \State \textit{// Extract historical data features}
    \State $H \leftarrow \text{Extract historical data features from } D$
    \State \textit{// Calculate technical indicators: $TE$}
    \State $T \leftarrow \text{Calculate $TE$ (EMA, MACD, RSI, etc.)}$
    \State \textit{// Combine features}
    \State $X \leftarrow H \cup T$
\EndProcedure

\Procedure{Model Training}{}
    \State \textit{// Initialize XGBoost regressor}
    \State $\mathcal{M} \leftarrow \text{XGBoost Regressor}$
    \State \textit{// Define hyperparameter grid}
    \State $G \leftarrow \{\eta, D_{\text{max}}, N, \lambda, \alpha, \gamma, S, C\}$
    \State \textit{// Perform grid search with cross-validation}
    \State $\theta \leftarrow \text{GridSearchCV}(\mathcal{M}, G, \text{scoring=RMSE})$
    \State \textit{// Train model with best hyperparameters}
    \State $\mathcal{M} \leftarrow \mathcal{M}.fit(X_{\text{train}}, Y_{\text{train}})$
\EndProcedure

\Procedure{Model Evaluation}{}
    \State \textit{// Predict on test set}
    \State $\hat{Y} \leftarrow \mathcal{M}.predict(X_{\text{test}})$
    \State \textit{// Calculate evaluation metrics}
    \State $\text{MAE} \leftarrow \frac{1}{n} \sum_{i=1}^{n} |y_i - \hat{y}_i|$
    \State $\text{RMSE} \leftarrow \sqrt{\frac{1}{n} \sum_{i=1}^{n} (y_i - \hat{y}_i)^2}$
    \State $R^2 \leftarrow 1 - \frac{\sum_{i=1}^{n} (y_i - \hat{y}_i)^2}{\sum_{i=1}^{n} (y_i - \bar{y})^2}$
    \State \textit{// Display results}
    \State \text{Display} (MAE, RMSE, $R^2$)
\EndProcedure

\State \textbf{return} Trained model $\mathcal{M}$

\end{algorithmic}
\end{algorithm*}

We obtained Bitcoin historical market data from Binance via the Binance API \cite{binance_data}. The dataset spans from February 1, 2021, to February 1, 2022, with a time interval of 15 minutes ($\Delta t = 15$ minutes). This interval was chosen for its balance between capturing detailed market fluctuations and maintaining accuracy. The data is split into 80\% for the training set and 20\% for the testing set. The choice of a shorter time interval is particularly important due to the high volatility of the Bitcoin market, where rapid changes are frequent. In such highly volatile markets, shorter intervals are essential for accurately capturing these swift price movements, unlike in less volatile markets where longer intervals might suffice.

Figure \ref{fig:bitcoin_close_price} illustrates the Bitcoin close price over time in USD. The x-axis represents dates, while the y-axis represents the price in USD. The plot provides a visual representation of the fluctuation in Bitcoin's closing price over the observed period, enabling insights into the cryptocurrency's price trend and volatility.

We take advantage of \texttt{StandardScaler} from \texttt{sklearn.preprocessing} module to scale the data. Let's denote the elements of the matrix \(X\) as \(x_{ij}\), where \(i\) represents the row index (sample) and \(j\) represents the column index (feature). The transformation applied by the \texttt{StandardScaler} to each feature \(j\) is outlined as follows:
\begin{enumerate}
    \item Compute the mean (\(\mu_j =\frac{1}{m} \sum_{i=1}^{m} x_{ij}\)) and standard deviation (\(\sigma_j =\sqrt{\frac{1}{m} \sum_{i=1}^{m} (x_{ij} - \mu_j)^2}\)) of feature \(j\), where \(m\) is the number of samples (rows), \(x_{ij}\) is the element at the \(i\)-th row and \(j\)-th column of \(X\).
    \item Apply the transformation to each element of feature \(j\): \[x'_{ij} = \frac{x_{ij} - \mu_j}{\sigma_j}\] where \(x'_{ij}\) is the scaled value of \(x_{ij}\).
\end{enumerate}

%1) Compute the mean (\(\mu_j =\frac{1}{m} \sum_{i=1}^{m} x_{ij}\)) and standard deviation (\(\sigma_j =\sqrt{\frac{1}{m} \sum_{i=1}^{m} (x_{ij} - \mu_j)^2}\)) of feature \(j\), where \(m\) is the number of samples (rows), \(x_{ij}\) is the element at the \(i\)-th row and \(j\)-th column of \(X\). 2) Apply the transformation to each element of feature \(j\): \[x'_{ij} = \frac{x_{ij} - \mu_j}{\sigma_j}\] where \(x'_{ij}\) is the scaled value of \(x_{ij}\).

%  --- Feature Engineering ---
\subsection{Feature Engineering} \label{subsec:feature_engineering}

%In this section, we elaborate on the various features incorporated in this case study. Specifically, we employ historical market data alongside with technical indicators as discussed next.

In this section, we elaborate on the various features incorporated in this case study, employing both historical market data and technical indicators.

\subsubsection{Historical Data} \label{subsec:historical_data}
In historical data analysis, we utilize various metrics to understand the behavior of Bitcoin prices within specific time periods. These metrics include:

\begin{itemize}
    \item Open price ($\mathcal{O}_p$): The initial price of Bitcoin at the beginning of a specific time period.
    \item Highest price ($\mathcal{H}_p$): The maximum price of Bitcoin recorded during a time period.
    \item Lowest price ($\mathcal{L}_p$): The minimum price of Bitcoin recorded during a time period.
    \item Close price ($\mathcal{C}_p$): The final price of Bitcoin at the end of a time period.
    \item Trading volume ($\mathcal{V}$): The total number of Bitcoin traded within a time period.
    \item Quote Asset Volume (QAV): The total trading value of Bitcoin within a time period.
    \item Number of Trades (NOT): The total number of trades executed during a time period.
    \item Total Buy Base Volume (TBBV): The total volume of Bitcoin bought during a time period.
    \item Total Buy Quote Volume (TBQV): The total value of Bitcoin bought during a time period.
\end{itemize}

\subsubsection{Technical Indicators}

Technical analysis indicators represent a trading discipline utilized to assess investments and pinpoint trading opportunities through the analysis of statistical trends derived from trading activities, including price movements and volume \cite{achelis2001technical}. In this study, we explore indicators to feed our machine learning model, such as EMA, MACD, relative strength index, momentum, price rate of change, and stochastic oscillator. 

%We employ Exponential Moving Average (EMA) with different periods; $\text{EMA}_{10}$, $\text{EMA}_{30}$, and $\text{EMA}_{200}$ represent the average price of Bitcoin over the last 10, 30, and 200 periods, respectively.

%We utilize Relative Strength Index (RSI) with different periods. RSI measures the magnitude of recent price changes to evaluate overbought or oversold conditions. Specifically, we use $\text{RSI}_{10}$, $\text{RSI}_{14}$, $\text{RSI}_{30}$, and $\text{RSI}_{200}$ measure the magnitude of recent price changes over 10, 14, 30, and 200 periods, respectively.

%We also employ Momentum (MOM) with different time periods to measure the rate of change in Bitcoin prices. In particular, $\text{MOM}_{10}$ and $\text{MOM}_{30}$ represent the rate of change in Bitcoin prices over the last 10 and 30 periods, respectively.

We employ EMA with different periods, where $\text{EMA}_{10}$, $\text{EMA}_{30}$, and $\text{EMA}_{200}$ represent the average price of Bitcoin over the last 10, 30, and 200 periods, respectively. To measure the magnitude of recent price changes and evaluate overbought or oversold conditions, we use RSI. Specifically, $\text{RSI}_{10}$, $\text{RSI}_{14}$, $\text{RSI}_{30}$, and $\text{RSI}_{200}$ assess price changes over 10, 14, 30, and 200 periods, respectively. In addition, we apply Momentum (MOM) indicators to gauge the rate of change in Bitcoin prices, with $\text{MOM}_{10}$ and $\text{MOM}_{30}$ reflecting changes over the last 10 and 30 periods, respectively.

%Furthermore, we incorporate Moving Average Convergence Divergence (MACD), a trend-following momentum indicator that shows the relationship between two moving averages of Bitcoin prices.

%Additionally, we use \%K10, \%K30, and \%K200 as components of the stochastic oscillator, comparing the current price of Bitcoin to its price range over the last 10, 30, and 200 periods, respectively.

%Finally, we involve the Percentage Rate of Change with 9 periods ($\text{PROC}_{9}$), which measures the percentage change in Bitcoin prices over the last 9 periods.

Furthermore, we incorporate MACD, a trend-following momentum indicator that illustrates the relationship between two moving averages of Bitcoin prices. Additionally, we use \%K10, \%K30, and \%K200 as components of the stochastic oscillator, which compare the current price of Bitcoin to its price range over the last 10, 30, and 200 periods, respectively. Finally, we include the Percentage Rate of Change with 9 periods ($\text{PROC}_{9}$), measuring the percentage change in Bitcoin prices over the last 9 periods.

%--- Methodology ---
\section{Methodology} \label{sec:mathematical_model}
This section details the proposed methodology for our machine learning approach to cryptocurrency price forecasting. Algorithm \ref{alg:crypto_forecasting} outlines our machine learning approach for cryptocurrency price forecasting using the XGBoost regressor model combined with various technical indicators such as EMA, MACD, RSI, and more. The process includes data collection and preprocessing, feature engineering, model training with hyperparameter tuning, and model evaluation. Details of this methodology are discussed in the following subsections.

Let $(x^{(i)}, y^{(i)})$ denotes a single sample/observation, and the set of samples is represented by:
\[
\mathcal{S} = \left\{(x^{(1)}, y^{(1)}), (x^{(2)}, y^{(2)}), \dots, (x^{(m)}, y^{(m)}) \right\}
\]
where \(x^{(i)} \in \mathbb{R}^{n}\) and \(y^{(i)} = \mathcal{C}_p^{(i)} \).

Considering both technical indicators and historical data for price prediction necessitates the integration of diverse datasets. To achieve this, we combine technical indicators and historical data as inputs to our model. The feature vector at a given time $t$ can be expressed as follows:

\begin{equation} 
\label{equation:feature}
\mathbf{x}^{(i)} =
\begin{bmatrix}
\mathcal{C}_p^{(i)} \\
\mathcal{V}^{(i)} \\
\text{QAV}^{(i)} \\
\text{NOT}^{(i)} \\
\text{TBBV}^{(i)} \\
\text{RSI}_{14}^{(i)} \\
\text{RSI}_{30}^{(i)} \\
\text{RSI}_{200}^{(i)} \\
\text{MOM}_{10}^{(i)} \\
\text{MOM}_{30}^{(i)} \\
\text{MACD}^{(i)} \\
\text{PROC}_{9}^{(i)} \\
\text{EMA}_{10}^{(i)} \\
\text{EMA}_{30}^{(i)} \\
\text{EMA}_{200}^{(i)} \\
\%K_{10}^{(i)} \\
\%K_{30}^{(i)} \\
\%K_{200}^{(i)} \\
\end{bmatrix}
, \quad \mathbf{x}^{(i)} \in \mathbb{R}^{n}
\end{equation}

To extend the generality of our model, we stack all feature vectors into a matrix $\mathbf{X}$, which can be expressed as follows:

\begin{equation}
\mathbf{X} =
\begin{bNiceMatrix}[first-row,
                    first-col,
                    last-col,
                    nullify-dots,
                    code-for-first-row = \color{blue},
                    code-for-first-col = \color{blue}
                    ]
& \mathcal{C}_p & \mathcal{V} &  \cdots & \%K_{200} & \\
  & x_{11} & x_{12} & \cdots &  x_{1m} &  \\
 & x_{21} & x_{22}  & \cdots & x_{2m} & \\
 & x_{31} & x_{32} & \cdots & x_{3m} &  \\
 & \vdots  & \vdots & \cdots  & \vdots & \\
 & x_{n-1 1} & x_{n-12} & \cdots & x_{n-1m} & \\
 & x_{n1} & x_{n2} & \cdots & x_{nm} & \\
\end{bNiceMatrix}
\end{equation}

Where:
\begin{align*} 
\mathcal{C}_{p} &= \begin{bmatrix} \mathcal{C}_{p}^{(1)} \\ \mathcal{C}_{p}^{(2)} \\ \vdots \\ \mathcal{C}_{p}^{(m)} \end{bmatrix},
\mathcal{V} &= \begin{bmatrix} \mathcal{V}^{(1)} \\ \mathcal{V}^{(2)} \\ \vdots \\ \mathcal{V}^{(m)} \end{bmatrix},
\hdots,
\%K_{200}  &= \begin{bmatrix} \%K_{200}^{(1)} \\ \%K_{200}^{(2)} \\ \vdots \\ \%K_{200}^{(m)} \end{bmatrix}
\end{align*}

The output matrix can then be expressed as follows:

\begin{equation*}
    Y = \begin{bmatrix} y^{(1)} \\ y^{(2)} \\ \vdots \\ y^{(m)} \end{bmatrix}
\end{equation*}

In this case study, the problem is to minimize the cost function for XGBoost regressor, which is a regularized finite-sum minimization problem defined as:
\begin{equation}
\min_{\Theta} J(\Theta) := \sum_{i=1}^{m_{\text{train}}} L(y_i, \hat{y}_i) + \sum_{k=1}^{K} \mathcal{R}(f_k)
\end{equation}

Where:
\begin{itemize}
  \item \( \Theta \) represents the set of parameters to be learned during training.
  \item \( L(y_i, \hat{y}_i) \) is the loss function that measures the difference between the true target value \( y_i \) and the predicted target balue \( \hat{y}_i \) for the \( i \)-th instance. In the context of this case study, we employ the mean squared error (MSE) loss function, which is expressed as follows:

  \begin{equation}
  L(y_i, \hat{y}_i) = \sum_{i=1}^{n} (y_i - \hat{y}_i)^2
  \end{equation}
  Here, \( y_i \) is the true target value for sample \( i \), and \( \hat{y}_i \) is the predicted target value for sample \( i \).

  \item \( \mathcal{R}(f_k) \) represents the regularization term for each tree to control its complexity. It typically includes both \( L_1 \) and \( L_2 \) regularization. Assuming \(T\) is the number of leaves in tree \(f_k\) and \(w_{j, k}\) is the weight for leaf \(j\) in tree \(f_k\), the regularization term for tree \(f_k\) is:
  \begin{equation}
      \mathcal{R}(f_k) = \gamma T + \frac{1}{2} \lambda \sum_{j=1}^T w_{j, k}^2 + \alpha \sum_{j=1}^T |w_{j, k}|
  \end{equation} 
  The regularization terms (\( \mathcal{R}(f_k) \)) help control the complexity of individual trees in the ensemble, preventing overfitting.
\end{itemize}

During training, XGBoost regressor aims to find the set of parameters (\( \Theta \)) that minimizes the overall cost function. The optimization is typically performed using techniques like gradient boosting, which involves iteratively adding weak learners to the ensemble to reduce the residual errors \cite{chen2016xgboost}.

\begin{table}[!t]
\caption{Parameter grid for GridSearchCV.}
\label{tab:param_grid}
\centering
\begin{tabular}{cc}
\toprule
\textbf{Parameter} & \textbf{Values} \\
\midrule
\( N \) & 300, 400 \\
\( \eta \) & 0.01, 0.1, 0.2 \\
\( D_{\max} \) & 3, 4 \\
\( W_{\text{min}} \) & 1, 3 \\
\( S \) & 0.8, 1.0 \\
\( C \) & 0.8, 1.0 \\
\( \gamma \) & 0, 0.1 \\
\( \alpha \) & 0.5, 1 \\
\( \lambda \) & 0.5, 1 \\
\bottomrule
\end{tabular}
\end{table}

Table \ref{tab:param_grid} presents a parameter grid used in GridSearchCV, a technique for hyperparameter tuning in machine learning models. Hyperparameters are predefined settings that control the learning process of algorithms. The table lists various hyperparameters commonly used in the XGBoost regressor model, a popular gradient boosting framework \cite{chen2016xgboost}. Each hyperparameter is accompanied by its corresponding values that are explored during the grid search process. For instance, \( N \) represents the number of estimators (trees) in the XGBoost model, with values of 300 and 400 being considered. Similarly, \( \eta \) denotes the learning rate, with potential values of 0.01, 0.1, and 0.2.

Other hyperparameters include \( D_{\text{max}} \) for maximum depth of trees, \( W_{\text{min}} \) for minimum child weight, \( S \) for subsampling ratio, \( C \) for column subsampling ratio, \( \gamma \) for minimum loss reduction required to make further splits, \( \alpha \) for L1 regularization term on weights, and \( \lambda \) for L2 regularization term on weights.

This parameter grid serves as a roadmap for systematically exploring various combinations of hyperparameters to identify the optimal configuration for the XGBoost model, thereby enhancing its predictive performance. The best combination of hyperparameters for the XGBoost model was selected based on the smallest RMSE, resulting in enhanced predictive performance. The chosen parameters are as follows:
\[
\Theta = \begin{pmatrix}
C & \gamma & \eta & D_{\text{max}} & W_{\text{min}} & N & \alpha & \lambda & S \\
0.8 & 0 & 0.2 & 4 & 3 & 300 & 1 & 0.5 & 1.0 \\
\end{pmatrix}
\]

Finally, the RMSE achieved with this parameter combination is the smallest observed during the hyperparameter tuning process.

% ---
% --- Results and Analysis ---
\section{Results and Analysis}\label{sec:results_and_analysis}

In this section, we provide simulations-based evaluations of the proposed machine learning model. In particular, we compute the Mean Absolute Error (MAE), RMSE, and R-squared ($R^2$).

\begin{equation}
\text{MAE} = \frac{1}{n} \sum_{i=1}^{n} |y_i - \hat{y}_i|
\end{equation}
MAE provides a simple and straightforward interpretation of the average absolute deviation between the predicted and actual values. It is easy to understand and is less sensitive to outliers compared to other metrics like RMSE.

\begin{equation}
\text{RMSE} = \sqrt{\frac{1}{n} \sum_{i=1}^{n} (y_i - \hat{y}_i)^2}
\end{equation}
RMSE provides a measure of the average magnitude of prediction errors in the same units as the target variable. It penalizes larger errors more heavily than MAE, making it particularly useful when large errors are undesirable.

\begin{equation}
R^2 = 1 - \frac{{\sum_{i=1}^{n} (y_i - \hat{y}_i)^2}}{{\sum_{i=1}^{n} (y_i - \bar{y})^2}}
\end{equation}

where $\bar{y}$ is the mean of the actual values of the target variable.

$R^2$ Score provides an indication of how well the model fits the data relative to a simple baseline model (e.g., a model that always predicts the mean). It ranges from 0 to 1, where higher values indicate a better fit. $R^2$ score is widely used for comparing different models and assessing overall model performance.

\begin{table}[h]
\centering
\begin{tabular}{l|c}
\hline
\textbf{Metric} & \textbf{Value} \\
\hline
RMSE & 59.9504 \\
MAE & 46.2229 \\
$R^2$ & 0.9999 \\
\hline
\end{tabular}
\caption{Model evaluation metrics.}
\label{tab:metrics}
\end{table}

Table \ref{tab:metrics} presents key evaluation metrics for our regression model. The RMSE is 59.9504, indicating the square root of the average squared difference between predicted and actual values. The MAE is 46.2229, indicating the average absolute difference between predicted and actual values. The model's $R^2$ Score is 0.9999, reflecting an exceptionally strong fit to the data. Overall, the model demonstrates high accuracy and predictive capability, with low errors and a near-perfect $R^2$ score.

Another way to assess the performance of the XGBoost Regressor model is to analyze the relationship between the predicted values and the residuals. Let \( y_{\text{test}} \) be the true target values from the test dataset, \( \hat{y}_{\text{pred}} \) be the predicted target values from the model, and \( \varepsilon \) be the residuals calculated as \( \varepsilon = y_{\text{test}} - \hat{y}_{\text{pred}} \).

% \begin{itemize}
%     \item \textbf{Residual Calculation}:
%     \[ \varepsilon = y_{\text{test}} - \hat{y}_{\text{pred}} \]
    
%     \item \textbf{Scatter Plot}:
%     The scatter plot represents the relationship between the predicted values \( \hat{y}_{\text{pred}} \) (x-axis) and the residuals \( \varepsilon \) (y-axis). Each point on the plot corresponds to a pair of predicted value and its corresponding residual.
    
%     \item \textbf{Horizontal Line at \( y = 0 \)}:
%     The horizontal line at \( y = 0 \) is added to the plot to denote the line where residuals equal zero. In an ideal scenario, the residuals should be evenly distributed around this line without showing any systematic pattern.
% \end{itemize}

\begin{figure}[!t]
   \centering
   \includegraphics[width=3.5in]{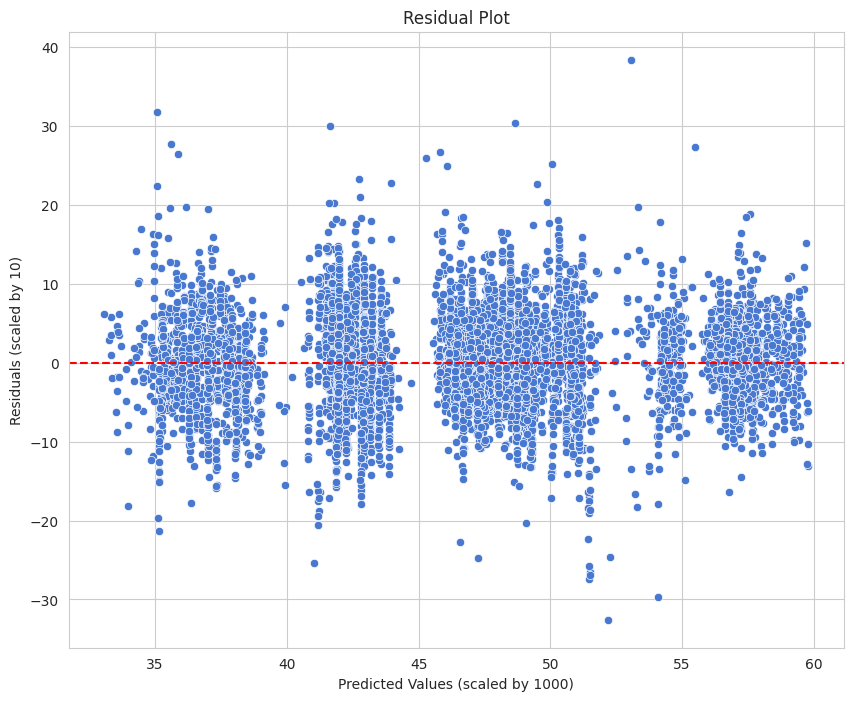}
   \caption{Scatter plot showing the residuals against the predicted values.}
   \label{fig:residual_plot}
\end{figure}

Figure \ref{fig:residual_plot} shows a scatter plot of the residuals against the predicted values. The plot displays the relationship between the predicted values (scaled by 1000) and the residuals (scaled by 10). A horizontal dashed line at \(y=0\) indicates perfect prediction, where residuals are centered around zero. The plot illustrates the model's ability to predict accurately across the range of predicted values.

\begin{figure}[!t]
   \centering
   \includegraphics[width=3.5in]{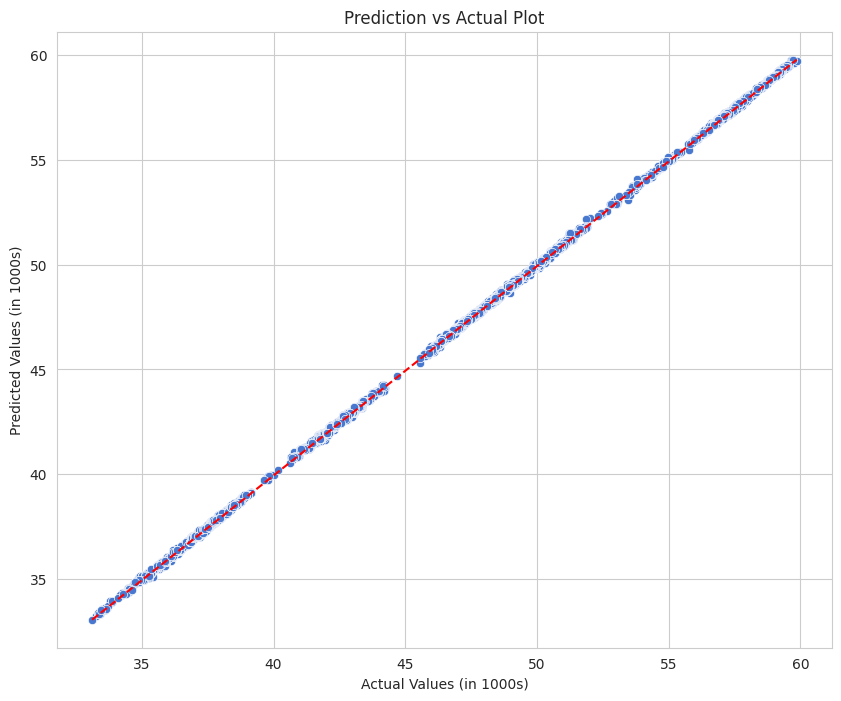}
   \caption{Scatter plot of actual vs. predicted values.}
   \label{fig:prediction_vs_actual}
\end{figure}

Furthermore, Figure \ref{fig:prediction_vs_actual} presents a scatter plot depicting the comparison between predicted values (in 1000s) and actual values (in 1000s). The diagonal dashed red line represents ideal prediction, where actual values align perfectly with predicted values. This plot offers insight into the model's efficacy across the spectrum of actual values, showcasing its predictive performance.

\subsection{State-of-the-Art Comparison} %\label{sec:discussion_and_comparison}
Lastly, this subsection provides a comparison between the work proposed in this paper and existing studies in the literature.

\begin{table*}[htbp]
\caption{Comparison of machine learning approaches in financial forecasting and trading.}
\label{tab:comparison}
\centering
\begin{tabular}{@{}p{2.4cm}p{2.5cm}p{3.7cm}p{3.2cm}p{3.8cm}@{}}
\toprule
\textbf{Paper} & \textbf{Methodology} & \textbf{Data Utilization} & \textbf{Model Performance} & \textbf{Contribution to Field} \\
\midrule
Shynkevich et al. \cite{shynkevich2017forecasting} & Machine learning algorithms & Daily price time series for stocks & Optimal performance with varied metrics & Impact of forecast horizon and input window length \\
\midrule
Liu et al. \cite{liu2021forecasting} & SDAE, deep learning & Historical data, public attention, macroeconomic environment & Superior performance in predictions & Improved prediction with SDAE, RMAE, and DA \\
\midrule
Jaquart et al. \cite{jaquart2022machine} & Ensemble machine learning models & Cryptocurrency market data & Statistically significant predictions & Long-short portfolio strategy \\
\midrule
Hafid et al. \cite{hafid2023bitcoin} & RF classifier & Historical data, few technical indicators & High accuracy in market trend prediction  & Effective market trend prediction (buy \& sell) \\
\midrule
Saad et al. \cite{saad2019toward} & Integration of economic theories with machine learning & User and network activity & High accuracy in price prediction & Understanding network dynamics \\
\midrule
Akyildirim et al. \cite{akyildirim2021prediction} & SVM, LR, ANN, RF & Historical price data, technical indicators & Predictive accuracy in price trends & Evidence of predictability in trends \\
\midrule
\textbf{This paper} & XGBoost regressor & Technical indicators, historical data & Low MAE, RMSE, $R^2 \approx 1$ & Novel machine learning strategy \\
\bottomrule
\end{tabular}
\end{table*}

%Table \ref{tab:comparison} shows a comparison of key studies focusing on financial forecasting/trading and this paper. It summarizes how each paper approaches the problem, what kind of data they use, how well their models perform, and what unique contributions they bring to the field. This comparison will help researchers and traders to understand the different methods and innovations in predicting financial trends and making trading decisions.

Table \ref{tab:comparison} provides a comprehensive comparison of various machine learning approaches in financial forecasting and trading. Shynkevich et al. \cite{shynkevich2017forecasting} leverage machine learning algorithms on daily stock price time series, achieving optimal performance by analyzing different forecast horizons and input window lengths. Similarly, Liu et al. \cite{liu2021forecasting} employ SDAE deep learning models utilizing historical data, public attention, and macroeconomic factors, which result in superior prediction accuracy. In addition, Jaquart et al. \cite{jaquart2022machine} implement ensemble machine learning models on cryptocurrency market data (streamed from CoinGecko \cite{coingecko}), producing statistically significant predictions and incorporating a long-short portfolio strategy. Furthermore, Hafid et al. \cite{hafid2023bitcoin} use a Random Forest classifier with historical data and a few technical indicators to achieve high accuracy in market trend prediction, effectively signaling buy and sell moments.

Saad et al. \cite{saad2019toward} integrate economic theories with machine learning, analyzing user and network activity to attain high accuracy in price prediction and offer insights into network dynamics. Moreover, Akyildirim et al. \cite{akyildirim2021prediction} apply SVM, LR, ANN, and RF algorithms on historical price data and technical indicators, demonstrating consistent predictive accuracy and trend predictability. In contrast, this paper introduces a novel approach using an XGBoost regressor with technical indicators and historical data, achieving low MAE, RMSE, and an $R^2$ value close to 1, thereby contributing a new machine learning strategy to the field.

% --- Conclusion ---
% ---
\section{Conclusion}\label{sec:conclusion} 
%Our experiments show that the XGBoost regressor offers valuable results, providing low errors and achieving near-perfect $R^{2}$ value.

%The proposed machine learning model takes into consideration numerous technical indicators (e.g., exponential moving average) and historical data (e.g., volume) as input, while providing an estimation of the close price. We fine-tune our model using a grid of parameters to select the best combination. Additionally, during the training process, we emphasize the regularization technique to avoid overfitting by assigning appropriate and non-null values to the regularization parameters (i.e., $\lambda$ and $\alpha$) in the cost function. Our findings contribute to advancing financial forecasting strategies that make use of machine learning models.

Our research highlights the efficacy of the XGBoost regressor model in forecasting Bitcoin prices using a combination of technical indicators and historical market data. The model's performance, as evidenced by the low Mean Absolute Error (MAE) and Root Mean Squared Error (RMSE) along with a near-perfect $R^{2}$ value, underscores its potential in providing accurate and reliable predictions in the highly volatile cryptocurrency market. By incorporating regularization techniques to mitigate overfitting and fine-tuning model parameters through an extensive grid search, we have achieved a robust predictive model. Furthermore, the use of various technical indicators such as the Exponential Moving Average (EMA), Moving Average Convergence Divergence (MACD), Relative Strength Index (RSI), and others, in conjunction with historical prices and volume data, has proven effective in enhancing the model's predictive capabilities. This approach not only offers a comprehensive analysis of market trends but also facilitates better decision-making for traders and investors.

This work contributes to the field of financial forecasting, particularly in the domain of cryptocurrency price prediction. The findings suggest that machine learning models, when properly calibrated and integrated with relevant technical indicators, can serve as powerful tools for navigating the complexities of financial markets. Future research could further explore the integration of additional data sources and advanced machine learning techniques to continue improving the accuracy and applicability of such models in dynamic trading environments.

%
% ---- Bibliography ----
%
\bibliographystyle{IEEEtran}
\bibliography{bibliography}
\end{document}